\documentclass[10pt,twocolumn,letterpaper]{article}

\usepackage{cvpr}              

\usepackage{graphicx}
\usepackage{amsmath}
\usepackage{amssymb}
\usepackage{booktabs}
\usepackage{verbatim}

\usepackage[pagebackref,breaklinks,colorlinks]{hyperref}

\usepackage[capitalize]{cleveref}
\crefname{section}{Sec.}{Secs.}
\Crefname{section}{Section}{Sections}
\Crefname{table}{Table}{Tables}
\crefname{table}{Tab.}{Tabs.}

\def\confName{CVPR}
\def\confYear{2022}

\begin{document}

\title{Precise Affordance Annotation for Egocentric Action Video Datasets}

\author{Zecheng Yu, Yifei Huang, Ryosuke Furuta, Takuma Yagi, Yusuke Goutsu, Yoichi Sato\\
Industrial Institute of Science, The University of Tokyo\\
{\tt\small \{zch-yu,hyf,furuta,tyagi,goutsu,ysato\}@iis.u-tokyo.ac.jp}
}
\maketitle

\begin{abstract}
   Object affordance is an important concept in human-object interaction, providing information on action possibilities based on human motor capacity and objects' physical property thus benefiting tasks such as action anticipation and robot imitation learning. However, existing datasets often: 1) mix up affordance with object functionality; 2) confuse affordance with goal-related action; and 3) ignore human motor capacity. This paper proposes an efficient annotation scheme to address these issues by combining goal-irrelevant motor actions and grasp types as affordance labels and introducing the concept of mechanical action to represent the action possibilities between two objects. We provide new annotations by applying this scheme on the EPIC-KITCHENS dataset and test our annotation with tasks such as affordance recognition. We qualitatively verify that models trained with our annotation can distinguish affordance and mechanical actions.
\end{abstract}

\section{Introduction}
\label{sec:intro}

Affordance was first defined by James Gibson~\cite{gibson1977concept} as the possibilities of action that objects or environments offer. It is a non-declarative knowledge we have learned for automatically activating afforded responses on an object decided by both our motor capacity, \textit{i.e.}, the motor actions suitable for human hands, and the object's physical properties such as shape. Recognizing affordance can benefit tasks like action anticipation and robot action planning by providing information about possible interactions with objects in the scene~\cite{liu2020forecasting}.

Many existing works~\cite{koppula2013learning, myers2015affordance, luddecke2017learning, nguyen2017object, thermos2017deep, nagarajan2019grounded} in computer vision investigated affordance. They use verbs as affordance labels to describe the possible actions associated with objects.
However, verbs like ``cut", ``take", and ``turn off" do not correspond the definition of affordance. More specifically: a) ``cut" is a possible action enabled by a knife, making the dataset fail to distinguish human natural motor capacities from the capacities extended by objects' functionalities; 
b) using ``take" as an affordance label overlooks changes in affordance when ``take" is performed with different grasp types, which cannot provide fine-grained affordance annotations;
and c) ``turn-off" is a goal-related action, but not a goal-irrelevant affordance. The affordance utilized in ``turn-off tap" should also apply in other interactions such as ``press button".

\begin{figure}[t]
    \centering
    \includegraphics[width=1.1\linewidth]{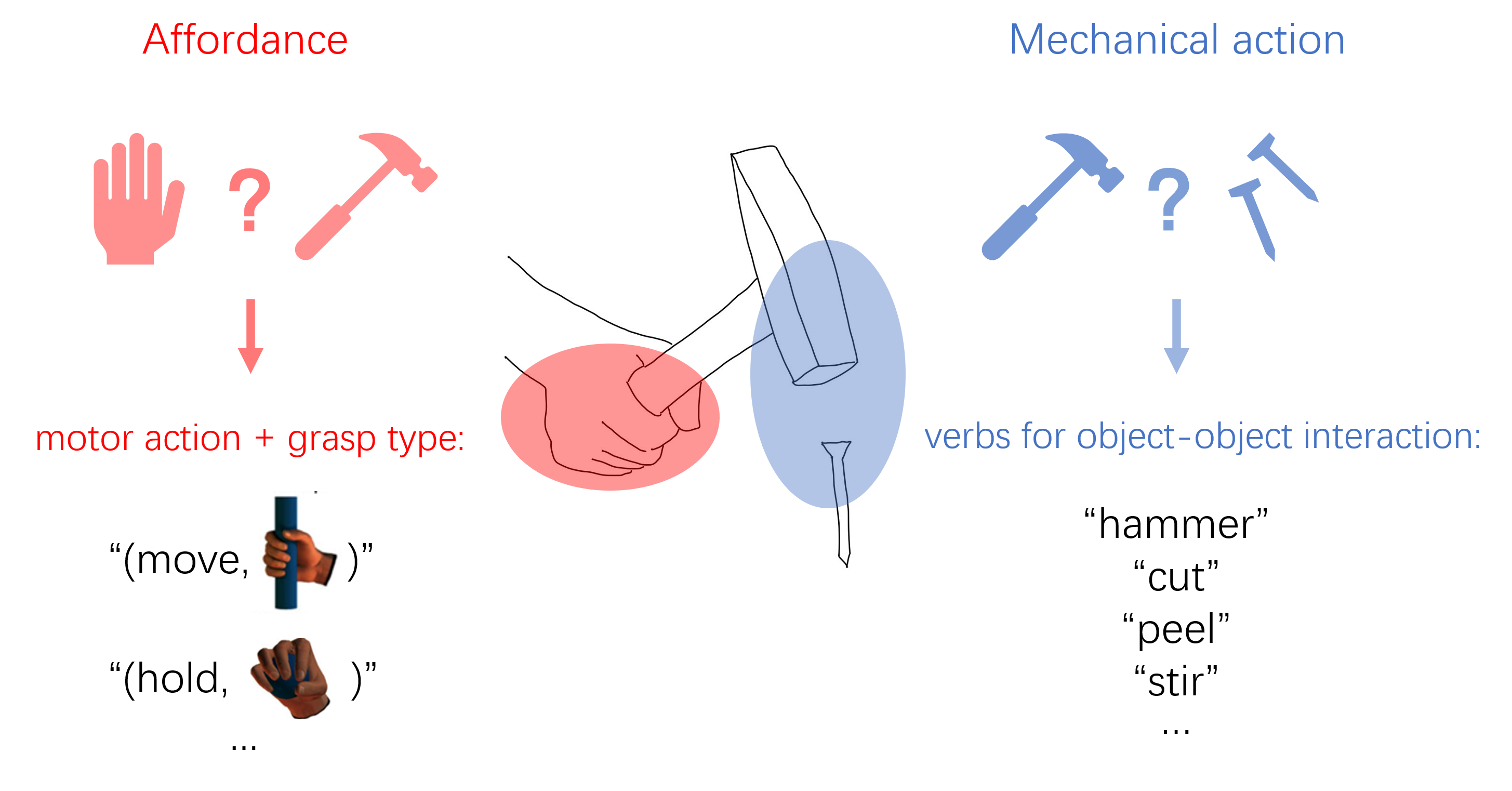}
    \caption{Definitions used in our annotation scheme: a) affordance labels are the combination of goal-irrelevant motor action and grasp type that exists in hand-centered interface; and b) mechanical action labels are action labels that can describe the interaction between two objects, exists in tool-centered interface (figures adapted from ~\cite{osiurak2017affordance}).}
    \label{fig:intro}
\end{figure}

In light of these issues, we need a precise affordance definition to distinguish affordance from other concepts, with the capability of representing human motor capacity. Inspired by the findings of neuroscience~\cite{osiurak2017affordance}, we address the shortcomings mentioned above by proposing an affordance annotation that considers hand-object interaction and tool-object interaction separately. Our annotation scheme: a) defines affordance as a combination of goal-irrelevant motor actions and hand grasp labels. This can represent the possible motor actions enabled by human motor capacity and object's physical property; b) defines mechanical actions as the possible actions between two objects, as shown in Figure~\ref{fig:intro}. 

Since annotating this information for a large-scale video dataset can be laborious, we propose an annotation method that leverages the consistency of affordance to simplify the annotation: the affordance will be the same when the same participant performs the same action on the same object. We believe that our newly proposed definition of affordance and the annotated dataset can facilitate a deeper understanding of object affordance and further improve subsequent tasks such as active object prediction, action anticipation, and robot imitation learning.

The main contributions of our work are as follows:
\begin{enumerate}
\item We point out the major shortcomings of existing affordance datasets, \textit{i.e.}, affordance is wrongly confused with object functionalities and goal-related actions. In addition, verbs cannot completely describe affordance because of neglecting the grasp types.
\item We propose a precise and efficient affordance annotation scheme to address issues above.
\item We provide annotations of affordance and other related concepts for a large-scale egocentric action video dataset: EPIC-KITCHENS~\cite{Damen2021RESCALING}. 
\end{enumerate}

\section{Related Works}
\label{sec:related}

\subsection{Affordance Datasets}

Earlier affordance datasets~\cite{nguyen2017object, myers2015affordance} annotated possible actions and the exact regions where actions could occur for object images. 
Koppula \textit{et al.}~\cite{koppula2013learning} provide affordance label annotation for human-object interaction video clips. Thermos \textit{et al.}~\cite{thermos2017deep} and Fang \textit{et al.}\cite{fang2018demo2vec} annotate human-object interaction hotspot maps as object affordance for video clips associated with their action labels. Furthermore, Nagarajan \textit{et al.}~\cite{nagarajan2019grounded} use the action labels of the EPIC-KITCHENS dataset, which is egocentric, as weak supervision to learn to generate human-object interaction hotspot maps. These datasets neither provide a clear definition of affordance nor consider humans' motor capacity. Therefore, we propose a precise affordance annotation scheme considering both humans' motor capacity and the object's physical property for video datasets.

\subsection{Affordance Understanding}

Affordance understanding methods can be divided into four categories: Affordance Recognition, Affordance Semantic Segmentation, Affordance Hotspots Prediction, and Affordance as Context. Given a set of images/videos, the task of affordance recognition~\cite{azuma_estimation_nodate} aims to estimate object affordance labels from them. Affordance semantic segmentation~\cite{luddecke2017learning, nagarajan2019grounded, fang2018demo2vec} aims at segmenting an image / video frame into a set of regions that are labeled with affordance labels. Some works ~\cite{liu2020forecasting,nagarajan2020ego} also use affordance as a context for other tasks such as action anticipating.

All of these methods are influenced by the lack of precise affordance annotation. For example, our attention differs when observing the ``cut'' action and the ``take'' action~\cite{huang2018predicting,huang2020mutual}: the former on the interacting object and the latter on the hand. Mixing them up may confuse models for affordance recognition. Moreover, depending on objects, we may perform the same action with different affordances. For example, we may directly push or handle the doorknob to close a door. However, previous works overlook these details by simply using the action "close" as an affordance label. This leads to the failure to distinguish different affordance hot spots in affordance semantic segmentation tasks.

\section{Proposed Affordance Annotation}
\label{sec:method}

Our goal is to develop a precise and efficient annotation scheme for affordance and other related concepts for egocentric datasets.

\subsection{Definitions}

Our proposed affordance annotation scheme is inspired by the three-action system model (3AS)~\cite{osiurak2017affordance}. The three-action system model includes affordance, mechanical action, and contextual relationship. We mainly focus on the first two concepts since they are closely related with hand-object interaction. Affordances are hand-centered, animal-relative, and goal-irrelevant properties of an object. Mechanical actions are tool-centered, and mechanical action possibilities between objects. Thus, we separately consider the affordance and mechanical action for an instance of hand-object interaction as shown in Figure ~\ref{fig:intro}. Affordance labels are defined as a combination of goal-irrelevant motor actions and grasp types, and we use verbs describing actions between objects as mechanical action labels.

To annotate an action video dataset, we first need to divide the original action labels of the dataset into tool-use actions and non-tool-use actions. Then annotate mechanical actions for tool-use actions and affordances for both tool-use and non-tool-use actions.

\subsection{Annotation Scheme}

\textbf{Tool-use / non-tool-use action annotation}: Tool-use / non-tool-use action annotation for action video datasets can be done by dividing original action labels of the dataset into three categories: tool-use action, non-tool-use action, and both, according to the meaning of each action label. For example, ``take” is a non-tool-use action, while ``cut" is a tool-use action. Some action labels could represent both the tool-use action and the non-tool-use action simultaneously, such as ``wash''. We ignore these labels during annotation because of their ambiguity.

\textbf{Mechanical action annotation}: We only need to annotate mechanical actions for tool-use actions. According to the definition, we can use the verbs of original action labels as mechanical action labels. For example: in ``stir food'', ``stir'' is the mechanical action between the slice and the food. We can automatically annotate mechanical actions for all tool-use action video clips based on this rule, allowing significant reduction of annotation cost.

\begin{figure}[t]
    \centering
    \includegraphics[width=1.0\linewidth]{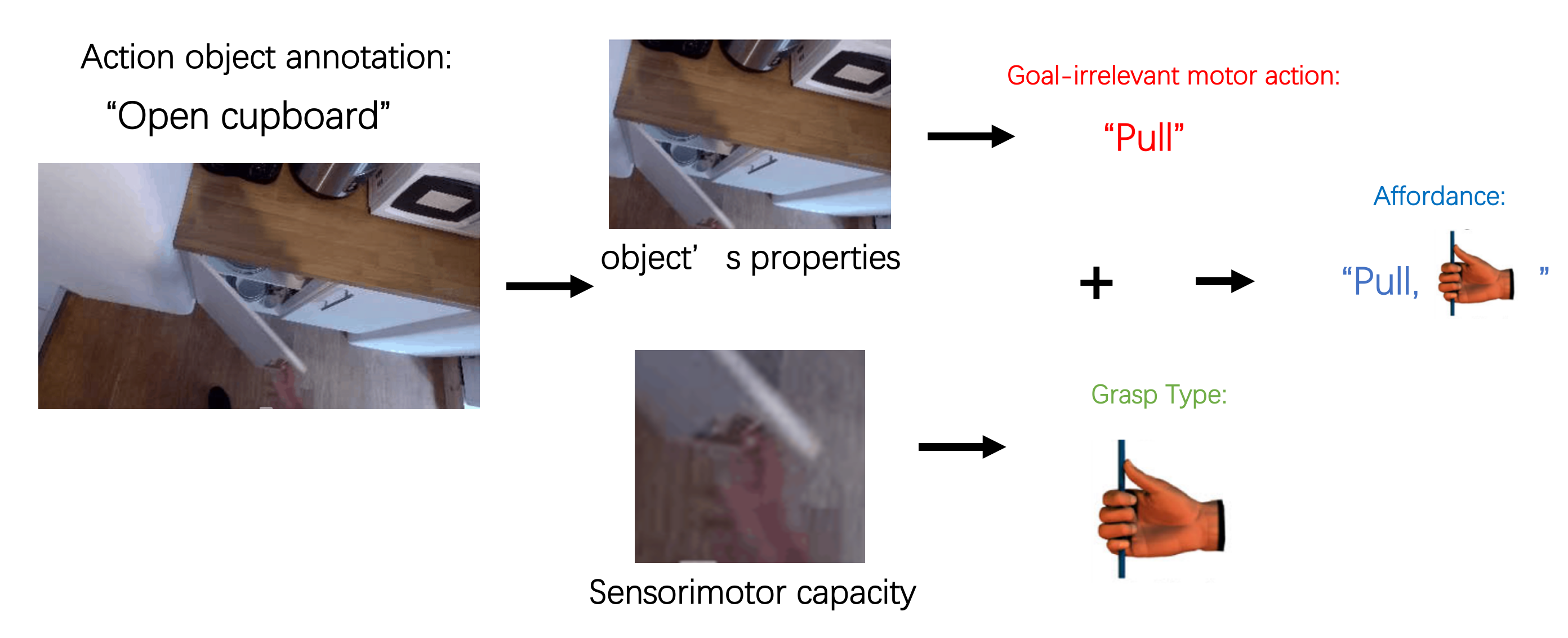}
    \caption{An affordance label is composed of a goal-irrelevant motor action label and a grasp type label.}
    \label{fig:affordancelabel}
\end{figure}

\textbf{Affordance annotation}: For affordance annotation, as shown in Figure~\ref{fig:affordancelabel}, we annotate a goal-irrelevant motor action and a grasp type for each video clip. Given an unlabeled video clip, we first define a goal-irrelevant motor action according to the object property used in it. In the example of this figure, we use ``pull'' to represent the ``pullable'' property of the cupboard. Next, we chose a grasp type from a 6-class grasp types taxonomy. This taxonomy is simplified from a well-known 33-class grasp type taxonomy~\cite{feix2015grasp} based on the power of the grasp type and the posture of the thumb. Finally, we combine the goal-irrelevant motor action label with the grasp-type label as the affordance label. This form of affordance labels can model both the object's physical property and the human motor capacity.

\begin{figure}[t]
    \centering
    \includegraphics[width=0.9\linewidth]{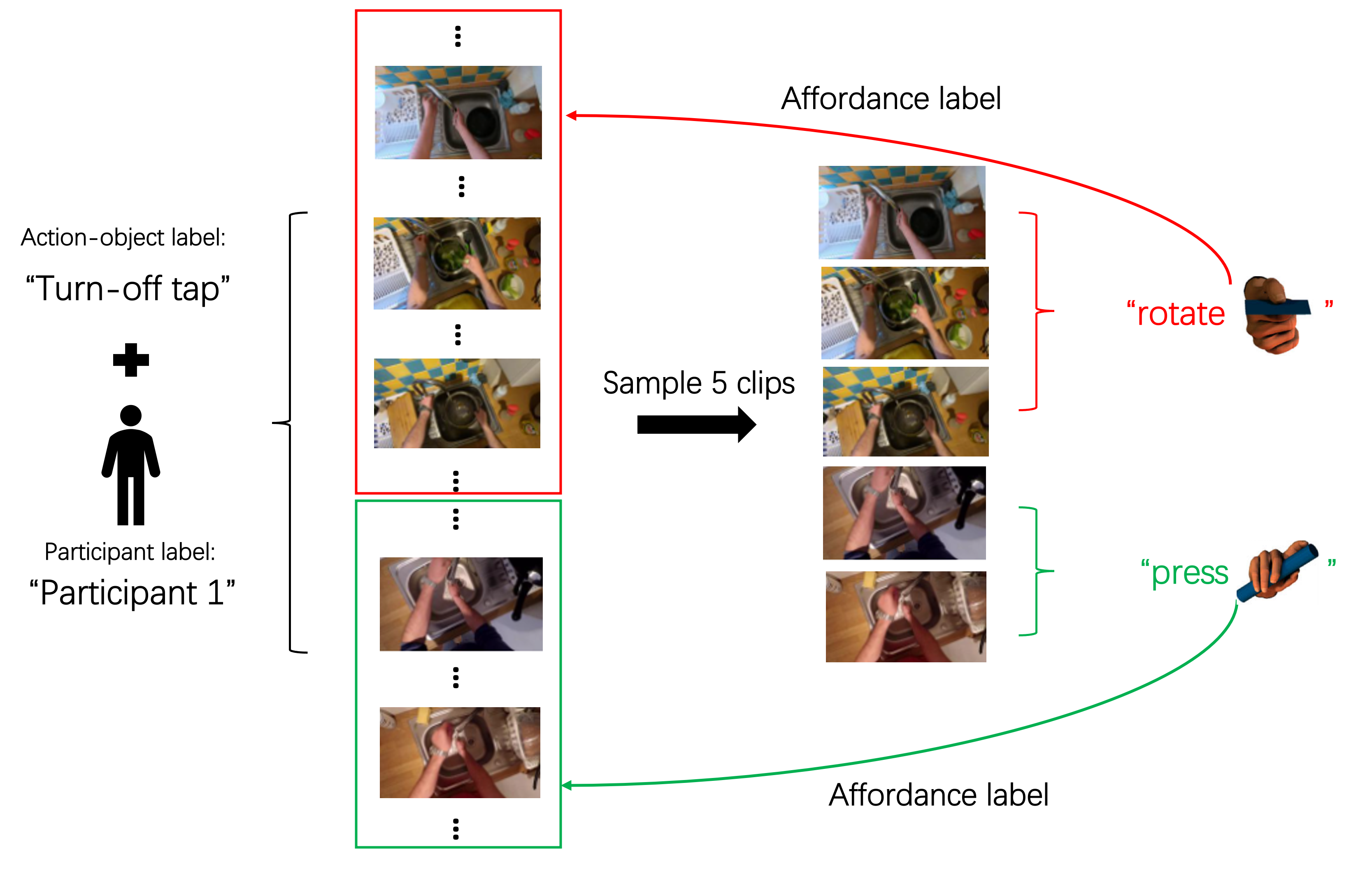}
    \caption{
    For efficient affordance annotation, we sample 5 video clips from the clips with the same verb(action)-noun(object)-participant, we use the annotation of these 5 clips as the annotation of all video clips in this group.}
    \label{fig:efficient}
\end{figure}

To reduce the manual work of affordance annotation, we propose an efficient annotation method. As shown in Figure~\ref{fig:efficient}, there are multiple video clips contain the same participant performing the same action on the same object in the original dataset. We first sample five video clips from video clips with the same verb (action)-noun (object)-participant annotation, then manually annotate affordances for them. Finally, we use these affordance labels as the affordance annotation for all video clips with the same verb (action)-noun (object)-participant annotation. The efficiency and accuracy of this method are shown in Section~\ref{sec:data}.

\section{Experiments}

We tested our annotation on three tasks to evaluate our proposed affordance annotation scheme.

\subsection{Dataset}
\label{sec:data}
We annotated the EPIC-KITCHENS dataset with our proposed method.
Inside the verb list of the original dataset, we select 33 verbs as mechanical action annotations. This results in 8.5k tool-use action video clips with mechanical action annotations. There are 51.5k non-tool-use action video clips in total. Then, using our annotation method, we sampled and annotated a total of 300 verb-noun pairs, obtaining 31,924 video clips and 24 affordance labels.
For quality check, we randomly sample 1,000 instances from the annotated video clips and manually checked their affordance label, getting an accuracy of 96.76\%.

\subsection{Tool-Use/Non-Tool-Use Action Classification Task}

To validate the rationality of our annotation, We compare the performance of a model trained with random annotation and a model trained with our annotation on the EPIC-KITCHENS dataset. We use a slowfast~\cite{feichtenhofer2019slowfast} model as the classifier. The result of tool-use/non-tool-use action classification is shown in Table~\ref{tabletoolnontool}. From the table, the model trained and tested with our annotation performs significantly better, while the model trained and tested on the random annotation can only get chance level performance.

\begin{table}[t]
  \centering
  \caption{Tool-use/non-tool use action classification results.}
  \scalebox{0.9}{
  \begin{tabular}{@{}lcc@{}}
    \toprule
    Dataset & Tool-use actions & Non-tool-use actions \\
    \midrule
    Random annotation & 0.4720 & 0.5282 \\ 
    Our annotation & 0.8580 & 0.7867\\ 
    \bottomrule
  \end{tabular}
  }
  \label{tabletoolnontool}
\end{table}

\subsection{Mechanical Action \& Affordance Recognition}

We benchmark the tasks of mechanical action recognition and affordance recognition. We also use slowfast for these two tasks. For the 33-class mechanical action recognition, we get a recognition accuracy of 51.90\%. For the 24-class affordance recognition task we obtain a recognition accuracy of 31.07\%.

\subsection{Visualization}

\begin{figure}[t]
    \centering
    \includegraphics[width=1.0\linewidth]{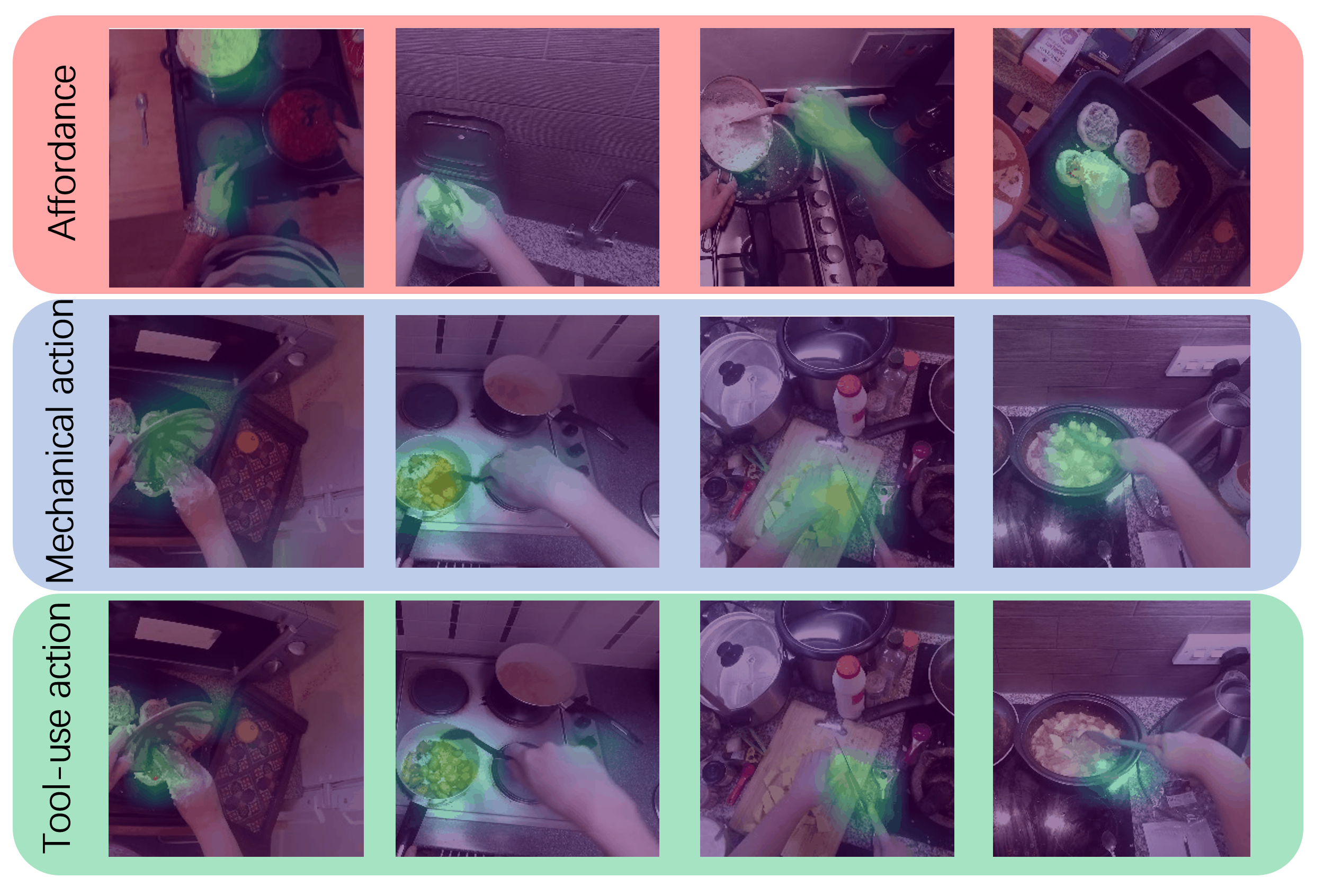}
    \caption{GradCam~\cite{selvaraju2017grad} visualization results of affordance recognition, mechanical action recognition, and tool-use/non-tool-use action recognition.}
    \label{fig:comparison}
\end{figure}

The visualization results generated by GradCam~\cite{selvaraju2017grad} are shown in Figure ~\ref{fig:comparison}. From the first row, we can see that the affordance recognition model focuses more on hands than on objects. The second row shows that the mechanical action recognition model cares more about the interaction between objects. When the second row is compared to the third row, it is evident that the mechanical action recognition model focuses on tool-object interactions (2nd row), and the tool-use/nontool-use action classification model focuses on the existence of tools (3rd row).

\section{Conclusion}

In this study, we proposed a precise affordance annotation scheme for egocentric action video datasets, which distinguishes affordance from other concepts like object functionality and reduces the manual annotation burden. We successfully applied our proposed annotation scheme to the EPIC-KITCHENS dataset and evaluated our proposed annotation on three tasks. The affordance recognition model focuses on hands, while the mechanical action recognition model focuses on the interaction between objects. This result shows that our affordance annotation completely represents affordance without missing human motor capacities and separates affordance from other related concepts. The results also confirm the assumption that affordance and mechanical actions are perceived differently, although they are all possible actions on objects. We believe that our proposed affordance annotation scheme can benefit tasks such as action anticipation and robot imitation learning by providing accurate information about hand-object interactions.

{\small
\bibliographystyle{ieee_fullname}
\bibliography{egbib}
}

\end{document}